\documentclass{article}
\usepackage[utf8]{inputenc}
\usepackage{spconf,amsmath,graphicx,xcolor,subfig,blindtext,CJKutf8,arydshln,hyperref}

\ninept

\title{Models of visually grounded speech signal pay attention to nouns: a bilingual experiment on English and Japanese}
\name{
	William N. Havard$^1$$^2$,  Jean-Pierre Chevrot$^2$, 
	Laurent Besacier$^1$
	\thanks{This work was supported by grants from NeuroCoG IDEX UGA as part of of the "Investissements d'avenir" program (ANR-15-IDEX-02)}
}

\address{
	$^1$LIG, Univ. Grenoble Alpes, CNRS, Grenoble INP, 38000 Grenoble, France\\
	$^2$LIDILEM, Univ. Grenoble Alpes, 38000 Grenoble, France
}
\begin{document}
%
\maketitle
\begin{abstract}
We investigate the behaviour of attention in neural models of visually grounded speech trained on two languages: English and Japanese. Experimental results show that attention focuses on nouns and this behaviour holds true for two very typologically different languages. We also draw parallels between artificial neural attention and human attention and show that neural attention focuses on word endings as it has been theorised for human attention.  Finally, we  investigate how two visually grounded monolingual models can be used to perform cross-lingual speech-to-speech retrieval. For both languages, the enriched bilingual (speech-image) corpora with  part-of-speech tags and forced alignments are distributed to the community for reproducible research.
\end{abstract}
\begin{keywords}
grounded language learning, attention mechanism, cross-lingual speech retrieval, recurrent neural networks.
\end{keywords}
\section{INTRODUCTION}
\label{sec:intro}


Over the past few years, there has been an increasing interest in research gathering the Language and Vision (LaVi) communities. Multimodal corpora such as Flickr30k \cite{young-etal-2014}  or MSCOCO \cite{MSCOCO} containing images along with natural language captions were made available for research. They were soon extended with speech modality: speech recordings for the captions of Flickr8k were collected by \cite{Harwath2015} \textit{via} crowdsourcing; spoken captions for MSCOCO were generated using Google Text-To-Speech (TTS) by \cite{Chrupala2017} and using Voxygen TTS by \cite{Havard2017}; extensions of these corpora to other languages than English, such as Japanese, were also introduced by \cite{Yoshikawa2017}.
These corpora, as well as deep learning models, lead to contributions in 
multilingual language grounding and learning of shared and multimodal representations with neural networks \cite{Chrupala2017, Alishahi2017, Harwath18_interlingua, harwath_raw_sensory, Gella2017, Kadar2017, HarwathG17,KamperSSL17}. 

This paper focuses on computational models of visually grounded speech that were introduced by \cite{Harwath2016UnsupervisedLO, Chrupala2017}. Learned representations of such models were analyzed by \cite{Kadar2017, Alishahi2017, Chrupala2017}: \cite{Kadar2017} introduced novel methods for interpreting the activation patterns of recurrent neural networks (RNN) in a model of visually grounded meaning representation from textual and visual input and showed that RNN pay attention to word tokens belonging to specific lexical categories. \cite{Chrupala2017} found that final layers tend to encode semantic information whereas lower layers tend to encode form-related information. \cite{Alishahi2017} showed that a non trivial amount of phonological information is preserved in higher layers, and suggested that the attention layer focuses on semantic information.

Such computational models can be used to emulate child language acquisition and could shed light on the inner cognitive processes at work in humans as suggested by \cite{DUPOUX18}. While \cite{Kadar2017, Alishahi2017, Chrupala2017} focused on analyzing  speech representations learnt by speech-image neural models from a phonological and semantic point of view, the present work focuses on lexical acquisition and the way speech utterances are segmented into lexical units and processed by a computational model of visually grounded speech. We analyze a key component of the neural model -- the attention mechanism -- and we observe its behaviour and draw parallels  between artificial neural attention and human attention. Attention indeed plays a key role in human perceptual learning, as stated by \cite{Gibson_1969}.


\textbf{Contributions.} 
We enrich an existing speech-image corpus in English with forced alignments and part-of-speech (POS) tags and analyse which parts of the spoken utterances the neural model attends to. In order to put these experiments in a cross-lingual perspective, we also experiment on a similar corpus in Japanese.\footnote{Both enriched corpora are available on \url{https://github.com/William-N-Havard/VGS-dataset-metadata}.} We show that the attention mechanism mostly focuses on nouns for both languages. We also show that our Japanese model developed a language-specific behaviour to detect relevant information by paying attention to particles, as Japanese toddlers do. Moreover, the bilingual corpus allows us to demonstrate that images can be used as pivots to automatically align spoken utterances in two different languages (English and Japanese) without using any transcripts. This preliminary result, in line with previous findings of \cite{Harwath18_interlingua}, confirms that neural speech-image models can capture a cross-lingual semantic signal, a first step in the perspective of learning speech-to-speech translation systems without text supervision.

    \begin{figure}[htbp]
        \centering
        \hspace{-0.5cm}
        \includegraphics[width=0.50\textwidth]{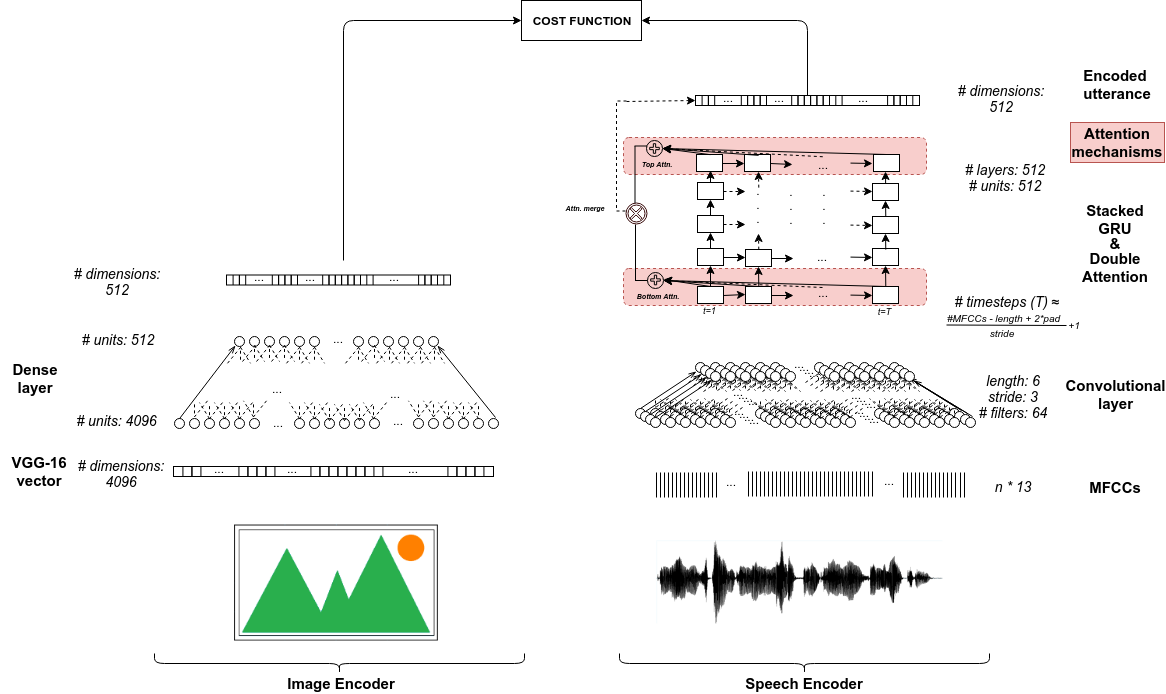}
        \caption{Neural model of visually grounded speech used in our experiments.}
        \label{fig:archi}
    \end{figure}

    \begin{figure*}[htbp]
    	    	\begin{minipage}{.40\linewidth}
    		\centering
    		\subfloat[]{\label{main:attn_en}\includegraphics[width=1\textwidth]{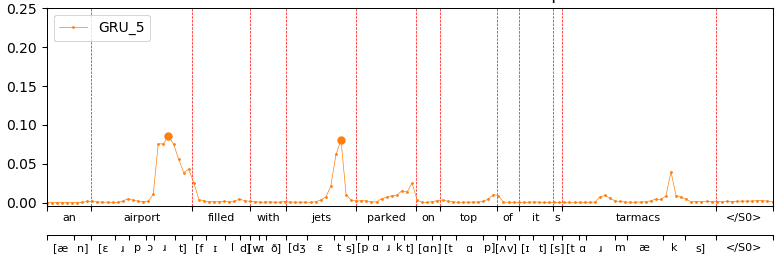}}
    	\end{minipage}
    	\hspace{-1cm}
    	\begin{minipage}{.3\linewidth}
    		\centering
    		\subfloat[]{\label{main:attn_img}\includegraphics[width=0.5\textwidth]{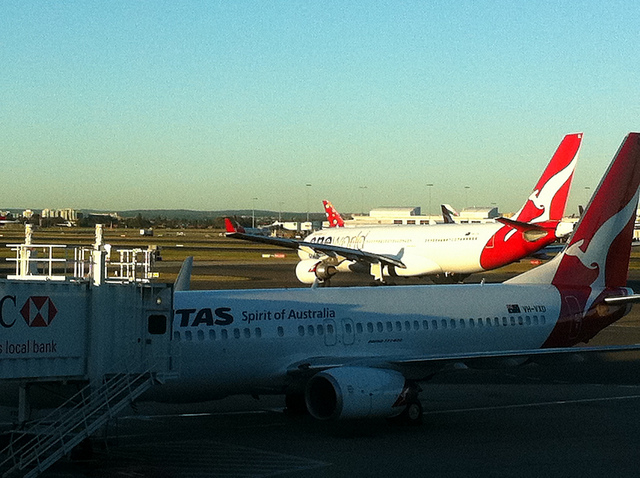}}
    	\end{minipage}
    	\hspace{-1cm}
    	\begin{minipage}{.40\linewidth}
    		\centering
    		\subfloat[]{\label{main:attn_jp}\includegraphics[width=1\textwidth]{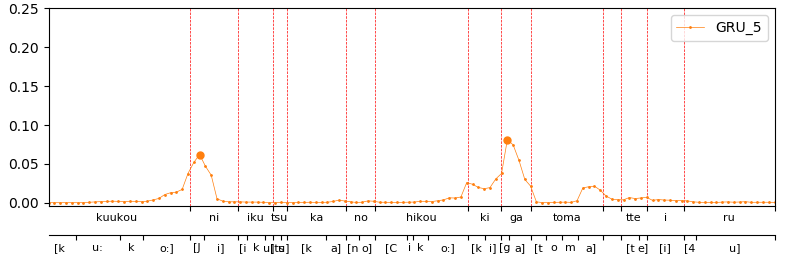}}
    	\end{minipage}
    	\caption{Attention weights  over an English (\ref{main:attn_en}) and Japanese caption (\ref{main:attn_jp}), both describing the same picture (\ref{main:attn_img}). Attention peaks in the English caption are located above ``\textsc{airport}" and ``\textsc{jets}". Attention peaks in the Japanese caption are located above ``\textsc{ni}" (particle indicating location) and ``\textsc{ga}" (particule indicating the subject of the sentence). Red dotted lines show token boundaries. Large orange markers show automatically detected peaks. Japanese caption reads: ``Several planes are stopped at the airport"}
    	\label{fig:attention_figures}
    \end{figure*}

\section{MODEL OF VISUALLY GROUNDED SPEECH}
\label{sec:majhead}
    The model we use for our experiments is based on that of \cite{Chrupala2017}. It is trained to solve an image retrieval task: given a spoken description it retrieves the closest image that matches the description. To do so, the model projects an image and its spoken description in a common representation space, so that matching image/utterance pairs lie near while mismatching image/utterance pairs lie apart.

        \subsection{General architecture}
        \label{subsec:general_architecture}
            The model (see figure \ref{fig:archi}) has two components: an image encoder, and a speech encoder. At training time, the network is presented with images and their corresponding spoken descriptions and tries to minimise the following loss function:

            
            \begin{equation}
                \resizebox{0.55\hsize}{!}{
                \begin{math}
                    \begin{split}
                        \sum_{u, i} 
                        \Bigg( \sum_{u'}\max [0, \alpha + d(u,i) - d(u',i)]\\
                        +\sum_{i'}\max[0, \alpha + d(u,i) - d(u,i')] \Bigg)
                    \end{split}
                \end{math}
                }
                \label{eq:loss_func}
            \end{equation}
            

            This loss function encourages the network to minimise by a margin $\alpha$ the distance $d(u,i)$ between the encoded image $i$ and the encoded utterance $u$ belonging to matching image/utterance pairs while making the distance greater for mismatching image/utterance pairs. 
    
        \subsection{Encoders}
        \label{subsec:modificiations}
            The image encoder takes VGG-16 (\cite{Simonyan14c}) pre-calculated vectors as input\footnote{VGG networks are trained to label images with a set of 1000 object categories from ImageNet.} instead of raw images. It only consists of a dense layer that learns how to shrink the 4096 dimensional VGG-16 input vector to a 512 dimensional vector, which is then L2 normalised. The speech encoder (input is 13 MFCC vectors instead of raw speech) consists of a convolutional layer followed by 5 stacked recurrent layers. Contrary to the original model (\cite{Chrupala2017}), we used GRU units instead of RHN units.\footnote{In fact, we aim at having a simpler model whose internal representations would be easier to understand as we also intend to study the gating mechanism in the future.} Results are still acceptable (see Table \ref{tab:results}) even if GRU architecture scores worse than original RHN one. 
            
        \subsection{Attention mechanism}
        \label{subsec:attention_mechanism}
            One of the key component of the model is its attention mechanism. 
            The model computes a weighted sum of the GRU activations at all timesteps as following: $\sum_{t}\alpha_t h_t\label{eq:attn}$. 
            Knowing by how much a given vector has been weighted gives us an insight on which portions of the speech signal the network relies to make its predictions (see Figure \ref{fig:attention_figures}). 
            In the original architecture (\cite{Chrupala2017}), attention follows the last recurrent layer. To have more insight on the representation learnt by the network, we added an attention mechanism after the first recurrent layer. Final vector produced by the speech encoder is a dot product of the vectors produced by both attentions. However, for the sake of clarity, we will only report in this paper results on the attention weights of the top attention mechanism \textsc{GRU5} (after the fifth recurrent layer).\footnote{Adding a second attention mechanism improves our results by -3 $\widetilde{r}$.}
            

\section{ENGLISH AND JAPANESE CORPORA}
\label{sec:corpora}
    The corpora we use for our experiments are based on MSCOCO \cite{MSCOCO}. MSCOCO is a dataset initially thought for computer vision purposes, mainly automatic image captioning. The dataset consists of a set of images, each paired with 5 written captions describing the image. All captions were written in English by humans and faithfully describe the content of the image. 
    The Japanese corpus we use is based on the newly created STAIR dataset \cite{Yoshikawa2017}. Using the same methodology as \cite{MSCOCO}, \cite{Yoshikawa2017} collected 5 Japanese captions for each image of the original MSCOCO dataset. As for the original MSCOCO dataset, Japanese captions were written by native Japanese speakers. It is worth insisting on the fact that these Japanese captions are original captions and not plain translations of their English equivalents. 
    MSCOCO and STAIR are thus comparable corpora.
    We trained our model on extended versions of MSCOCO and STAIR. Spoken COCO dataset was introduced by \cite{Chrupala2017} for English.
    We followed the same methodology as \cite{Chrupala2017} and generated synthetic speech for each caption in the Japanese STAIR dataset. We created the spoken  STAIR dataset so it would follow the exact same train/val/test\footnote{566 435, 25 000, and 25 000 captions in each set respectively.} split as \cite{Chrupala2017}. 
    We thus have two comparable corpora: one featuring images and spoken captions in English, and another one featuring the same images and spoken captions in Japanese. This allowed us to compare the behaviour of the same architecture on two typologically different languages.
                
    \begin{table}[htbp]
      \centering
        \begin{tabular}{c|c|c|c|c}
          \textbf{Model} & \textbf{R@1} & \textbf{R@5} & \textbf{R@10} & \textbf{$\widetilde{r}$}\\\hline
          English  & 0.060 & 0.195 & 0.301 & 25\\\hline
          Japanese & 0.054 & 0.180 & 0.283 & 28\\
        \end{tabular}
      \caption{
        Recall at 1, 5, and 10 results as well as median rank \textbf{$\widetilde{r}$} on a speech-image retrieval task (test part of our datasets with 5k images). Original implementation by \cite{Chrupala2017} with RHN reports median rank \textbf{$\widetilde{r}=13$} on  English dataset. Chance for median rank  $\widetilde{r}$ is 2500.5.
        }
      \label{tab:results}
    \end{table}

    We forced aligned each spoken caption to its transcription (using the Montreal Forced Aligner \cite{MontrealFA} and Maus Forced Aligner \cite{kisler2017multilingual} for English and Japanese respectively), resulting in alignments at word and phone level. We also tagged each dataset using TreeTagger \cite{schmid:treetagger}
    for English and KyTea \cite{kytea}
    for Japanese. As the tagset of both taggers differs, we mapped each POS to its Universal POS equivalent \cite{PETROV12.274}
    enabling us to compare the POS distribution of each corpus.\footnote{We decided to map KyTea's \textsc{TAIL} tags -- word conjugation -- to Universal \textsc{VERB} tag, thus the high proportion of verbs in the Japanese dataset.}
    
\section{WHAT DO MODELS PAY ATTENTION TO?}
\label{sec:analysis_of_attention}
   We first train two monolingual models for English and Japanese on the train set (566 435 spoken captions) of the corpora for 15 epochs.
   Baseline results are similar for English and Japanese (see Table \ref{tab:results}).
   
   To analyse the behaviour of the attention mechanism of our model,
   we encoded each caption of the test set and extracted the attention weights \begin{math}\alpha_t\end{math},
   resulting in an array of \begin{math}t\end{math} weights. We then used a peak detection algorithm\footnote{Uses the first order difference of the input array - see \url{https://github.com/lucashn/peakutils}.} to detect local maxima in the attention weights and thus know which timesteps were given the highest weights (large orange markers in Fig. \ref{fig:attention_figures}). We only considered peaks that were at least 60\% as high as the highest detected peak in the utterance.
    
        \begin{table}[htbp]
            \resizebox{0.5\textwidth}{!}{
                \begin{tabular}{c:c:c|cc:c:c}
                \multicolumn{3}{c}{English}         & \multicolumn{4}{c}{Japanese}                                \\
                word       & peak freq. & ref. freq & word         & gloss               & peak freq. & ref. freq \\\hline
                toilet     & 2.16       & 0.17      & ga           & subject part.       & 17.83      & 5.25      \\
                baseball   & 1.84       & 0.22      & no           & topic part.         & 9.53       & 6.24      \\
                train      & 1.71       & 0.25      & o            & direct object part. & 6.6        & 0.59      \\
                giraffe    & 1.6        & 0.11      & ni           & location part.      & 6.55       & 3.58      \\
                skateboard & 1.57       & 0.14      & de           & location part.      & 1.81       & 1.72      \\
                sign       & 1.33       & 0.19      & piza         & ``pizza"             & 1.47       & 0.13      \\
                kitchen    & 1.17       & 0.18      & to           & ``with" part.        & 1.04       & 1.37      \\
                with       & 1.13       & 2.09      & ke:ki        & ``cake"              & 1.02       & 0.1       \\
                frisbee    & 1.11       & 0.11      & shimauma     & ``zebra"             & 0.99       & 0.09      \\
                cake       & 1.03       & 0.11      & suke:tobo:do & ``skateboard"        & 0.98       & 0.13     
                \end{tabular}
            }
            \caption{Top 10 focused words for English and Japanese. ``Peak freq." refers the number of attention peaks (in \%) above a given word. ``ref. freq." refers to the frequency of the same word token in the training set.}
            \label{tab:examples_words_en_jp}
        \end{table}
    \vspace{-1cm}
    \subsection{Which morpho-syntactic categories are highlighted by attention?}
    \label{subsec:behaviour_of_attention}
\begin{figure}[htbp!]
        	\begin{minipage}{\linewidth}
        		\centering
        		\subfloat[]{\label{main:pos_training}\includegraphics[width=0.96\textwidth]{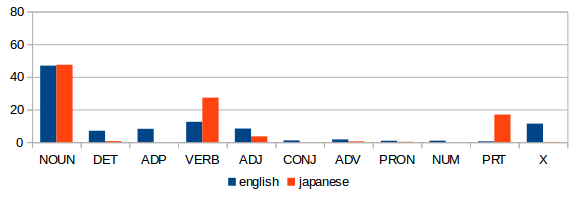}}
        	\end{minipage}
        	\begin{minipage}{\linewidth}
        		\centering
        		\subfloat[]{\label{main:pos_attn}\includegraphics[width=0.96\textwidth]{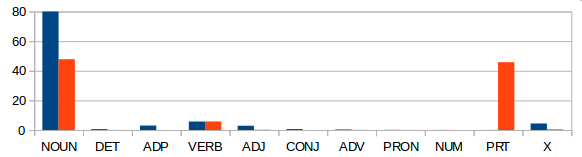}}
        	\end{minipage}
        	\caption{(a) Baseline POS distribution if attention peaks were to occur randomly. (b) POS distribution of words under detected attention peaks. English (blue) and Japanese (red).}
        	\label{fig:attention_pos}
        \end{figure}

    Having a timestep aligned speech signal for each language enables us to see above which words (and thus POS) attention focuses on. Table \ref{tab:examples_words_en_jp} shows the top ten words located under peaks for both languages (and their corresponding frequency in the training corpus). In order to see if the attention mechanism does any better than learning corpus statistics, we need a baseline POS distribution for comparison. One possibility would be to simply compare the proportion of peaks under a given POS to the frequency of the same POS computed on tokens (as provided in Table \ref{tab:examples_words_en_jp}). However, by doing so, we would assume that all tokens have the same length in the speech signal, which is not the case (verbs are longer than determiners for instance). 
    Thus, for each spoken utterance of the test set, we sampled $50*p$ random peak positions ($p$ number of true detected peaks per utterance), and computed the POS distribution over such peaks (see \ref{main:pos_training}). We consider this as our baseline corpus distribution if attention peaks were to occur randomly.
    
        \vspace{-0.27cm}
        \subsubsection{English}
        We notice (Fig. \ref{main:pos_attn}) that the attention mechanism of the English model primarily focuses on \textsc{NOUNS}: 82\% of the peaks are located above nouns. This is far above corpus frequency, which is 47\%. The attention mechanism considers neither determiners (\textsc{DET}) nor adpositions (\textsc{ADP}) nor adjectives (\textsc{ADJ}) as relevant as only 0.6\%, 3\%, and 2.85\% are highlighted, where corpus frequencies would predict 7\%, 8\%, and 8\% respectively. Verbs (\textsc{VERB}) are half as often highlighted as corpus frequency would predict, meaning attention barely relies on such words to make its prediction.
        
        \vspace{-0.27cm}
        \subsubsection{Japanese}
        The Japanese attention mechanism clearly makes use of particles\footnote{Particles are small suffixed grammatical words.} (\textsc{PRT}): 45.77\% of the peaks are located above such words where corpus frequency would predict 16.9\%. In fact, 6 of the top ten words are particles (see Table \ref{tab:examples_words_en_jp}). Moreover, 17.83\% of the peak highlight speech segments corresponding to the \textsc{GA} particle, well before nouns: \textsc{GA} is a particle that is used to indicate that the preceding word is the subject of the sentence. Thus, detecting such a particle is most useful, as the preceding word surely is the main object of the target image. 
        The Japanese attention mechanism also seems to rely on nouns as 47.79\% of peaks are located above nouns. One could argue this value is not very different from corpus frequency: 47.42\%. However, if such POS were to hinder prediction, we would expect the attention mechanism to lower the number of peaks above such words, such as the model did for verbs or adjectives, which is not the case here, meaning \textsc{NOUNS} are useful for the model's prediction.
        
        \vspace{-0.27cm}
        \subsubsection{Child language acquisition and noun-bias}
        When learning their native language, it has been theorised that children exhibit a noun-bias \cite{Gentner_noun_bias}:\footnote{\cite{Gentner_noun_bias} states that ``words that refer to concepts are easy to learn because the child has already formed object concepts, and need only match words and concepts".} that is, in most languages children learn nouns before any other catagory. We notice that both models exhibit such language-general behaviour and favour nouns over other categories. Also, we showed that our Japanese model develops a language-specific behaviour when mainly focusing on \textsc{GA} particles. \cite{Haryu2016} demonstrated that Japanese toddlers also make use of \textsc{GA} to segment speech before any other particle. The \textit{noun-bias} phenomenon in our corpus can be explained by two factors: first, images in our corpus display many objects, thus prompting annotator to use more nouns than verbs; second, VGG vectors (used to encode images) are only trained to detect objects and not actions.
    
        \subsection{Attention above word beginnings or word endings?}
        \label{subsec:peaks_pos}
        \begin{table}[htbp]
        \centering
                \begin{tabular}{c|c:c:c:c}
                             & Beginning & Middle-Beg. & Middle-End & End \\\hline
                    EN  & 6.19      & 9.14        & 39.24      & \textbf{45.42} \\
                    JA & 27.90     & 18.70       & 17.58      & \textbf{35.80}
                \end{tabular}
            \caption{Position of attention peaks above words for English (EN) and Japanese (JA).}
            \label{tab:location}
        \end{table}
        
        We analysed above which part of words peaks are located. We divided each word beneath a peak into 4 equal parts and counted the percentage of peaks located above a given category (see Table \ref{tab:location}). We notice that peaks in our English model are mainly located on the second half of the words. This phenomenon is coherent with Slobin's \cite{slobin} Operating Principles favoring language acquisition stating that children \textit{``pay attention to the ends of words"}. Peaks in Japanese are located at word endings but also at word beginnings. It seems the very beginning of some particles is able to trigger an attention peak. 


\section{IMAGES AS PIVOTS FOR CROSS-LINGUAL SPEECH RETRIEVAL?}

We have seen in previous section that  attention focuses on nouns and Table \ref{tab:examples_words_en_jp} suggests that these nouns correspond to the main concept of the paired image. To confirm this trend, we experiment on a cross-lingual speech-to-speech retrieval task using images as pivots. 
 
This possibility was introduced in  \cite{Harwath18_interlingua}, 
but required training jointly or alternatively two speech encoders within the same architecture and a parallel bilingual speech dataset 
while we experiment with separately trained models for both languages. In \cite{Harwath18_interlingua}, a parallel corpus was needed as the loss functions adopted try to minimise either the distance between captions in two languages or the distance between captions in two languages and the associated image as pivot. As our approach uses two monolingual models, we do not need a parallel corpus. Each monolingual model can be trained on its own dataset featuring images and their spoken description.
The approach is the following: we first select a set of pivot images never seen by any of the monolingual models before. 
We encode these images with the image encoder of each language.\footnote{Since both image encoders (from English and Japanese) are trained separately, they do not  lead to the same representation of an image.} Then, for each speech utterance query in a source language $u_{src}$ (English for instance), we find the nearest speech utterance in the target language $u_{tgt}$ (Japanese for instance) which minimises the cumulated distance $d(u_{src},i)+d(i,u_{tgt})$ among all pivot images $i$.

    \begin{table}[htbp!]
      \centering
      \resizebox{0.35\textwidth}{!}{
            \begin{tabular}{c|c|c|c|c}
              \textbf{Query} & \textbf{R@1} & \textbf{R@5} & \textbf{R@10} & \textbf{$\widetilde{r}$}\\\hline
              EN $\rightarrow$ JP  & 0.087 & 0.327 & 0.519 & 9.94\\\hline
              JP $\rightarrow$ EN  & 0.087 & 0.326 & 0.521 & 9.84\\\hline\hline
              \cite{Harwath18_interlingua} EN $\rightarrow$ HI  & 0.034 & 0.114 & 0.182 & --\\\hline 
              \cite{Harwath18_interlingua} HI $\rightarrow$ EN  & 0.033 & 0.121 & 0.203 & --\\\hline
            \end{tabular}
        }
      \caption{
        Results on English (EN) to Japanese (JP) and Japanese to English speech-to-speech retrieval (subset of 1k captions). For comparison,  we report \cite{Harwath18_interlingua}'s results on English to Hindi (HI) and Hindi to English  speech-to-speech retrieval. Chance scores are R@1=.001, R@5=.005, and R@10=.01. Chance  for median rank  $\widetilde{r}$ is 500.5.
        }
      \label{tab:results_multilingual_retrieval}
    \end{table}

To make sure no parallel dataset is used, we trained a new English model on the first half of the train set, and a new Japanese model on the second half. We evaluated our approach on 1k captions of our test corpus to be comparable with \cite{Harwath18_interlingua}.\footnote{We did not perform evaluation on the full $25000_{EN} \times 25000_{JP}$ distance matrices where each source query is associated with 5 target captions. Instead, we randomly sub-sampled ten $1000_{EN} \times 1000_{JP}$ distance matrices so that there would be only one target caption for each query in order to compare our results with \cite{Harwath18_interlingua}. Results are averaged over 10 random samples.}
At the time of the evaluation, given a speech query in language $src$ which we know is paired with image $I$, we assess the ability of our approach to rank the matching spoken caption in language $tgt$ paired with image $I$ in the top 1, 5, and 10 results and give its median rank $\widetilde{r}$.
We report our results in Table \ref{tab:results_multilingual_retrieval} as well as results from \cite{Harwath18_interlingua} who performed speech-to-speech retrieval using crowd-sourced spoken captions in English and Hindi.
\\Our results are surprisingly high given the fact we did not train a bilingual model but used the output of two monolingual models never trained to solve such a task. Nevertheless, it is also important to mention that \cite{Harwath18_interlingua} experimented on real speech with multiple speakers while we used synthetic speech with only one voice.
Table \ref{tab:multilingual_query} shows an example of top-1 retrieved Japanese sentences for 2 English queries.


\vspace{-0.15cm}
\begin{table}[htbp!]
    \centering
        \begin{tabular}{ll}
            EN          & this is a display of donuts on a couple shelves                           \\
            JA          & \begin{CJK}{UTF8}{min}いろいろな種類のドーナツが並べられている\end{CJK}   \\
            Trans.      & Different kinds of donuts are lined up                                    \\\hline
            EN          & a living room with some brick walls and a fireplace                       \\
            JA          & \begin{CJK}{UTF8}{min}ソファーやテーブルや暖炉のある西洋風の部屋\end{CJK} \\
            Trans.      & Western-style room with sofa, table and fireplace                         \\
       \end{tabular}
    \caption{Example of semantically related captions. English (EN) query and retrieved Japanese caption (JA) and its translation (TRANS).}
    \label{tab:multilingual_query}
\end{table}

\vspace{-0.75cm}
\section{CONCLUSION}

In this paper we showed that attention in a neural model of visually grounded speech mainly focuses on nouns. We also showed that this behaviour holds true for two very typologically different languages such as English and Japanese and that attention could also develop language-specifc mechanisms to detect relevant information in one of the languages. We also provided evidence that it is possible to perform speech-to-speech retrieval with images as pivots using the output of two independently trained monolingual models. In future work, we would like to validate our methodology on a bilingual dataset featuring real voices and try to extract a bilingual speech-to-speech dictionary using attention peaks as anchor points. \\Ultimatly, we would like to emphasise the paramount importance of using other languages than English when trying to analyse the linguistic representations learnt by neural networks so as to understand if the models encode language specific or language general information, and thus better understand their strengths and weaknesses.
\label{sec:conclusion}

\vspace{-0.25cm}
\section{ACKNOWLEDGEMENTS}
\label{sec:acknowledgements}
\vspace{-0.25cm}
We thank G. Chrupa{\l}a and his team for sharing their code and dataset, as well as for helping us with technical issues.

\bibliographystyle{IEEEbib}
\bibliography{strings}

\end{document}